\documentclass{article}

 \usepackage[preprint]{neurips_2026}

\usepackage[utf8]{inputenc} 
\usepackage[T1]{fontenc}    
\usepackage{hyperref}       
\usepackage{url}            
\usepackage{booktabs}       
\usepackage{amsfonts}       
\usepackage{nicefrac}       
\usepackage{microtype}      
\usepackage{xcolor}         

\usepackage{amsmath}
\usepackage{amssymb}
\usepackage{mathtools}

\usepackage[capitalize,noabbrev]{cleveref}
\usepackage{amsthm}

\theoremstyle{plain}
\newtheorem{theorem}{Theorem}[section]

\theoremstyle{definition}
\newtheorem{definition}[theorem]{Definition}

\newtheorem{remark}[theorem]{Remark}
\newtheorem{observation}{Observation}[section]
\newtheorem{fact}{Fact}[section]

\usepackage[textsize=tiny]{todonotes}

\usepackage[utf8]{inputenc} 
\usepackage[T1]{fontenc}    
\usepackage{url}            
\usepackage{booktabs}       
\usepackage{amsfonts}       
\usepackage{nicefrac}       
\usepackage{microtype}      
\usepackage{xcolor}         
\usepackage{amsfonts}
\usepackage{bm}
\usepackage{enumitem,array,booktabs}
\usepackage{algorithm}
\usepackage{algorithmic}
\newcounter{protocol}

\usepackage{framed}
\usepackage{natbib}
\usepackage{bbm}
\usepackage{tabulary,multirow}
\usepackage{makecell}
\usepackage{dsfont}
\usepackage{tcolorbox}
\usepackage{ulem}
\usepackage{thmtools}
\usepackage{xpatch}
\usepackage{tablefootnote}

\newcommand{\1}{\mathds{1}}

\newcommand{\E}{\mathbb{E}}

\newcommand{\cA}{\mathcal{A}}

\newcommand{\cD}{\mathcal{D}}

\newcommand{\cH}{\mathcal{H}}

\newcommand{\cL}{\mathcal{L}}

\newcommand{\cS}{\mathcal{S}}

\newcommand{\cX}{\mathcal{X}}

\newcommand{\cY}{\mathcal{Y}}

\renewcommand{\epsilon}{\varepsilon}

\newcommand{\nothere}[1]{}

\usepackage{color-edits}
\addauthor{lc}{blue}
\addauthor{ym}{red}
\usepackage{graphicx}  
\usepackage{wrapfig}   
\usepackage{thm-restate}
\allowdisplaybreaks

\usepackage{cleveref}

\usepackage[table]{xcolor}

\title{Monoculture Robust Learning}

\author{%
  Lee Cohen\\
  Stanford University \\
  \And
  Jon Kleinberg \\
  Cornell University \\
  \And
    Omer Reingold \\
  Stanford University\\ 
  }

\begin{document}

\maketitle

\begin{abstract}
We introduce a notion of monoculture robustness in Probably Approximately Correct (PAC) learning, as a way to mitigate the problem of \textit{algorithmic monoculture} \citep{KleinbergR21Monoculutre}: the tendency of models sharing the same components (e.g., training data) to fail on the same inputs. Specifically, we ask whether a learner with access to a single dataset can output $k$ individually accurate hypotheses whose joint failure probability is comparable to training on $k$ independent datasets. We say that such a learning procedure achieves \textit{monoculture robustness}, and we quantify this by bounding the \textit{conjunctive error} that all $k$ hypotheses misclassify points in an individual sense. 
We design methods for achieving this type of guarantee by drawing a surprising connection between replicability~\citep{Impagliazzo22} and monoculture robustness. Specifically, we show that if the base learner is $\rho$-replicable, then running it independently $k$ times on the same sample yields individual monoculture error that shrinks exponentially with $k$, controlled by the single-run error of the base learner at the point and $\rho$. We complement these positive results with a lower bound: for individual monoculture robustness, 
we prove a worst-case lower bound showing that, for learners, driving the individual conjunctive error down to $\gamma$ can require a sample-size multiplier of $t = \Omega(\log(1/\gamma))$.
\end{abstract}

\section{Introduction} 
In many real-world settings, people find themselves in situations where $k$ competing agents (or firms) offer the same type of opportunity or benefit, and each agent uses a learned decision rule to decide whether to extend an offer to a given individual. The individual may receive offers from multiple agents but ultimately accepts at most one. Examples include loan approvals across multiple banks, screening across multiple employers, ticket access for the same concert through different agencies, discounted travel offers during the holidays across competing platforms, or promotional discounts (e.g., a discounted haircut) across several salons in the same promotion network. If these agents rely on the same model, e.g., a single ad system that serves the ads for all the agents, their mistakes can become correlated, and the same individual can be rejected by all $k$ agents, even though the environment appears to provide multiple ``chances.''

This is the problem of \textit{algorithmic monoculture} \citep{KleinbergR21Monoculutre}, in which the use of a common underlying model by multiple competing firms leads to correlated errors and consequently adverse consequences for individuals (and sometimes for the firms themselves). It is a common concern with modern machine learning systems, which are often deployed in settings where multiple providers use a common training pipeline and/or common dataset, and therefore their algorithms may \textit{repeatedly misclassify the same inputs}. This means that even if an individual appears to be getting $k$ ``tries'' at an opportunity when they apply to $k$ different providers, they might in fact be getting something that more resembles a single try at the opportunity, since the misclassifications of all $k$ providers' models are very similar.


In this paper, we seek strategies to mitigate the negative effects of algorithmic monoculture, asking: if we only have access to one dataset, can we still produce $k$ accurate hypotheses whose \textit{joint} failure probability is significantly lower than a single hypothesis? We formalize this question in the Probably Approximately Correct (PAC) setting by introducing \textit{individual monoculture robustness} and its average analogue, \textit{expected monoculture robustness}. We consider a learner that, given a single sample $S \sim \mathcal{D}^n$, outputs $k$ hypotheses $h_1,\ldots,h_k$, and measure monoculture failure via the \textit{conjunctive error}: the event that \textit{all} $k$ hypotheses misclassify a fixed point $x$ or, in expectation, a random point $x \sim \mathcal{D}_X$. This captures the ``shared blind spot'' that motivates monoculture concerns: beyond standard accuracy guarantees, it cares about the \textit{correlation} of errors across the returned hypotheses, not just their marginal risks.

This has a direct connection to \textit{algorithmic fairness}~\citep{Dwork12Fairness}. Fairness and accuracy guarantees are typically stated for a single predictor, but in the settings above an individual effectively faces $k$ predictors in parallel (or over short time horizons), and what matters is the chance that all of them reject the individual. When errors are correlated across agents, “apply to more places” or “try another place” provides far less protection than one would expect under independence, and the harm can be concentrated on the same people or groups across many decision-makers. Monoculture robustness is designed to capture this concern by shifting attention from marginal error of each model to the joint failure probability across the $k$ models.

A notable conceptual point is that monoculture failure can arise from two sources of randomness. 
Sometimes the error is \textit{sample-driven}: for a point $x$, most internal random seeds fail whenever the drawn sample is unrepresentative for $x$. In this regime, rerunning the same algorithm on the same sample will not help much, and reducing the conjunctive error requires genuinely more data. Other times the error is \textit{internal randomness driven}: for a typical sample $S$, there are many seeds that predict the correct label, so rerunning with fresh randomness can reduce the probability that \textit{all} $k$ runs land in the ``bad'' seeds. This separation guides our lower bound and the positive results. 
Throughout this work, we study both fundamental limits on monoculture robustness and concrete procedures for achieving it, including ERM-based and replicable learners.
\subsection*{Contributions.}
Motivated by practical monoculture failures and the desire to ``simulate'' independence from a single dataset, we outline our contributions as follows.

\textbf{1. Problem formulation.} In section~\ref{sec:model}, we formalize monoculture robustness for PAC learning by modeling a learner that, given a single dataset, outputs $k$ accurate hypotheses intended to serve $k$ agents. We quantify monoculture failure via conjunctive error and distinguish expected monoculture error from the stronger individual  version. We also clarify when individual guarantees are meaningful, namely for a point $x$ with nontrivial single-run error of the base learning algorithm $q_x$ (it is inherently uninformative to demand small joint failure when the single-run error is large).

\textbf{2. Lower bounds and unavoidable sample increase}. 
We show that, in the realizable setting, individual monoculture robustness cannot in general be improved beyond what additional independent data provides: there exist simple classes and an input $x$ for which driving the individual monoculture error at $x$ down to $\gamma$ can require a sample-size multiplier $t=\Omega(\log(1/\gamma))$. In particular, for sample-driven failures, this sample-size increase is generally unavoidable.

\textbf{3. Replicability and same-sample joint error reduction}. In section~\ref{sec:replicability} we analyze (pointwise) replicability~\citep{Impagliazzo22,hopkins2025approximate} as a black-box condition under which rerunning a learner independently $k$ times on the same sample provably helps: the individual monoculture error shrinks exponentially with $k$, controlled by $q_x$ and the replicability parameter.  This section also makes two conceptual points: (i) Replicability might appear counterintuitive for achieving monoculture robustness. After all,  replicability is a stability property where a learning algorithm produces w.h.p. \textit{the exact same hypothesis} when trained on different samples, provided the internal randomness seed remains fixed. We show that it can still produce diversity across seeds on a fixed sample; and (ii) monoculture robustness is weaker than replicability, so while replicability provides individual monoculture robustness guarantees ``for free'', it is useful only if we require replicability regardless.

\textbf{4. Expected Monoculture Error. }
When relaxing the goal to minimize the expected monoculture error, we consider a simple sample-splitting strategy from~\cite{Aden-AliHLZ24} in Appendix~\ref{sec:expected}. We observe that in the realizable proper learning case, a result from~\cite{Aden-AliHLZ24} implies that splitting the sample $S$ of size $2n$ into two independent halves and running an ERM on each yields a convergence rate of $O(d/n)$, where $d$ is the VC dimension of the hypothesis class. This improves upon the standard expected error lower bound of $\Omega\left(\frac{d}{n} \log \frac{n}{d}\right)$ for a single ERM \citep{HAUSSLER1994248}, effectively ``gaining'' a logarithmic factor. It demonstrates a real separation between duplicating a single hypothesis trained on all the data and producing independent hypotheses by splitting the same data budget. 
This observation also highlights that if we do not require replicability and merely want low expected monoculture error for a given data budget, subsampling/splitting can be the right tool.

For brevity, omitted proofs are relegated to Appendix~\ref{sec:proofs}. 
\vspace{-2mm}
\subsection{Related Work}
The phenomenon of algorithmic monoculture was formalized by~\citet{KleinbergR21Monoculutre}, who demonstrated that while shared scoring rules may appear locally optimal for decision-makers, they introduce correlated failures that can reduce overall social welfare compared to noisy, independent evaluators. \citet{Bommasani22Homogenization} provided crucial empirical grounding for this theory, introducing the metric of outcome homogenization to quantify the extent to which distinct systems, specifically those relying on shared pre-trained components, systematically exclude the same individuals. This work supports the ``component sharing hypothesis'', suggesting that the widespread adoption of the same foundation model or models trained on the same data by different agents exacerbates systemic exclusion.

Complementing these findings, \citet{Black22Multiplicity} and \citet{Marx20PredictiveMultiplicity} explore the "Rashomon Effect" or model multiplicity, highlighting that for many tasks, there exists a multiplicity of equally accurate models that nonetheless yield conflicting predictions for individual subjects. While multiplicity implies that diversity is theoretically achievable, standard training pipelines often fail to realize it without explicit intervention. Extending monoculture analysis to strategic settings, \citet{Peng24matchingmarkets} examined two-sided matching markets, showing that while monoculture can help applicants identify viable matches, it disproportionately increases outcome variance for high-risk individuals. Unlike these works, which primarily characterize the existence and societal impact of correlated errors, our paper focuses on the learning-theoretic limits of overcoming them. 

Our work is also related to self-consistency \citet{Cooper23Self-Consistency}, which measures agreement between predicted labels across training samples without reference to the true label, while our goal is to control the label-dependent joint error of individually accurate predictors.

We provide a detailed discussion and 
additional related work on self-consistency, model multiplicity, predictive multiplicity, algorithmic fairness, and replicability in Appendix~\ref{app:related}.

\vspace{-3mm}
\section{Model}\label{sec:model}

We focus on the binary classification task in the Probably Approximately Correct (PAC) learning setting. 
Let $\cX$ be the input space, and $\cY = \{0,1\}$ the label space.
Let  $\cH \subset \cY^\cX $ represent a hypothesis class with finite Vapnik-Chervonenkis (VC) dimension, $d$. Consider a distribution $\cD$ over $\cX \times \cY$, and let $f^*: \cX \to \cY$ be the target classifier. In the realizable setting, we assume $f^* \in \cH$. 
An Empirical Risk Minimization (ERM) algorithm takes as input a sample $S = \{(x_i, f^*(x_i))\}_{i=1}^n$ and returns a hypothesis $h \in \cH$. While the  goal is to minimize the population loss (the true error rate),
$
\cL_D(h) = \mathbb{E}_{\cD}[\1(h(x) \neq f^*(x)],
$
an ERM algorithm works by minimizing the loss on the provided sample, $S$. 
In the realizable case, an ERM algorithm returns a hypothesis $h \in \cH$ that is consistent with the sample.
The returned hypothesis minimizes the empirical loss on $S$, where the empirical loss is
$
\cL_S(h) = \frac{1}{n} \sum_{i=1}^n \1[h(x_i) \neq f^*(x_i)].
$
To accommodate algorithms that output an ensemble of hypotheses, we extend the definition of (realizable or agnostic) PAC learnability to the setting where an algorithm returns $k$ hypotheses, requiring that standard learning guarantees hold for each.

\begin{definition}[PAC Learnability]\label{def:PAC}
A hypothesis class $\cH$ is PAC learnable in the agnostic setting (or realizable, if $\min_{h \in \cH} \cL_\cD(h) = 0$) if there exists a learning algorithm $\cA$ and a sample complexity function $n_{\cH}(\epsilon, \delta)$, such that for any $\epsilon, \delta \in (0, 1)$ and any distribution $\cD$ over $\cX \times \cY$, the following holds. When $\cA$ runs on a sample $S \sim \cD^n$ of size $n \ge n_{\cH}(\epsilon, \delta)$, it returns hypotheses $h_1,\ldots,h_k$ for some $k\in \mathbb{N}^+$ such that for every $i\in[k]$,
\[
\Pr_{S \sim \cD^n} \left[ \cL_\cD(h_i) \le \inf_{h \in \cH} \cL_\cD(h) + \epsilon \right] \ge 1 - \delta.
\]
In this case $\cA$ is an $(\epsilon,\delta)$-learner for $\cH$.
\end{definition}

We remark that optimal sample complexity bounds are known for any hypothesis class $\cH$ with finite VC dimension $d$. Specifically, $n_{\cH}(\epsilon, \delta) = \Theta\left(\frac{d + \log(1/\delta)}{\epsilon}\right)$ in the realizable setting, and $n_{\cH}(\epsilon, \delta) = \Theta\left(\frac{d + \log(1/\delta)}{\epsilon^2}\right)$ in the agnostic setting (see~\citet{Shalev-Shwartz_Ben-David_2014} and~\citet{Hanneke16}).

We quantify monoculture failure via the conjunctive error, which measures the simultaneous misclassification by all hypotheses.
\vspace{-3mm}
\paragraph{Conjunctive error.} Suppose algorithm maps  samples into a $k$-tuple of hypotheses $\textbf{h}=(h_1,\ldots,h_k)^T$. 
Fix any $x \in \cX$. We define the event \textit{$k$-conjunctive error} of $\cA$ on $x$ as
\[
e_k(\textbf{h},x)=\1 \left[\wedge_{i=1}^k (h_{i}(x)\ne f^*(x))\right].
\]
Next, we define monoculture robustness and individual monoculture robustness as algorithmic properties of learners.
\begin{definition}[Expected Monoculture-Robustness]
An algorithm $\cA$ is a $(\gamma,\epsilon,\delta)$-\textit{expected monoculture-robust}  if 
$\cA$ is an $(\epsilon,\delta)$-learner 
and 
\[
\E_{x \sim \cD_\cX, S\sim \cD^n, \cA} \left[ e_k(\textbf{h},x) = 1 \right]\leq \gamma.
\]
We call $\E_{x \sim \cD_\cX, S \sim \cD^n\cA} \left[ e_k(\textbf{h},x) = 1 \right]$ the \textit{expected  monoculture error} of $\cA$.
\end{definition}

Namely, a $(\gamma,\epsilon,\delta)$-expected monoculture robust algorithm is an $(\epsilon,\delta)$-learner with small expected monoculture error. Appendix~\cref{sec:expected} interprets this expected-error objective through a sample-splitting result of~\citep{Aden-AliHLZ24}, showing how sample splitting can improve the expected monoculture error in the realizable proper setting.
\begin{definition}[Individual Monoculture-Robustness]
An algorithm $\cA$ is $(\gamma_x,\epsilon,\delta)$-\textit{individual-monoculture-robust } for $x\in \cX$ if 
$\cA$ is an $(\epsilon,\delta)$-learner 
and 
\[
\Pr_{S\sim \cD^n, \cA} \left[ e_k(\textbf{h},x) = 1 \right]\leq \gamma_x.
\]
We call $\Pr_{S\sim \cD^n, \cA} \left[ e_k(\textbf{h},x) = 1 \right]$ the \textit{individual monoculture error} of $\cA$ on $x$.
\end{definition}
That is, an algorithm is $(\gamma_x,\epsilon,\delta)$-individual-monoculture-robust for $x$ if it is an $(\epsilon,\delta)$-learner with small individual monoculture error at $x$.

We note that if for every $x\in\cX$ an algorithm satisfies  $(\gamma_x,\epsilon,\delta)$-individual monoculture robustness, it is also $(\E_{x\sim D_\cX}[\gamma_x],\epsilon,\delta)$-expected monoculture robust.

For the special case of $k=1$, we denote the \textit{single-run error} of $\cA$ on $x$ as
\begin{align*}
q_x:&=q_x(\cA,n,D)=\Pr_{S \sim \cD^{n},\cA}[e_1]=\Pr_{S\sim \cD^n,\cA}[f^*(x)\ne (\cA(S))(x)].  
\end{align*}
That is, $q_x$ is the probability that $\cA$ errs on $x$ when trained on a random sample of size $n$ and returns a single hypothesis. 

\paragraph{Sample Partitioning Meta Algorithm. }
A special type of algorithm is the following: given a random sample $S$ of size $tn$, take an ERM algorithm $A$, and run it on an arbitrary or random partition of $S$ into $t\leq k$ independent samples $S_1,\ldots,S_t$. 
If $k>t$, it returns the $k$-tuple $\textbf{h}=(A(S_1),\ldots,A(S_t),A(S_1),\ldots,A(S_1))^T$. 

Compared to $q_x$, the individual monoculture error of a sample partitioning meta algorithm for each $x\in\cX$ decays exponentially with $t$. 
\begin{observation}
For every $x\in \cX$, a sample partitioning meta algorithm holds
\[
\Pr_{S_1,\ldots,S_t\sim \cD^n}[e_k(\textbf{h},x)]=q_x^t.
\]
\end{observation}

Moreover, we use this type of algorithm as a baseline for individual monoculture robustness with $n=n_\cH(\epsilon,\delta)$ and $t=k$. The rationale is that training each agent on an independent sample ensures that their prediction errors on any given individual are mutually independent.

\textbf{Goal.} 
Given parameters $\epsilon, \delta \in (0,1)$ and $k \in \mathbb{N}^+$, and a sample size $n = n_{\mathcal{H}}(\epsilon, \delta)$, our objective is to design a meta-algorithm that uses a base $(\epsilon, \delta)$-learner and is: 

\textbf{- Expected Monoculture Robust. } Here, our goal is to minimize the expected monoculture error. Specifically, we aim for $\gamma$ to outperform the standard ERM expected error.

\textbf{- Individual Monoculture Robust.} For any $x \in \mathcal{X}$ with a bounded single-run error (e.g., $q_x<1/2$), we seek to minimize the individual monoculture error.\footnote{Since we use existing algorithms as black boxes (e.g., ERM), it is only meaningful to ask for small joint failure at points where a single run is already has non-zero probability to be correct; if the single-run error is $1$, no procedure can make the “all $k$ fail” event reliably small without additional information.} We aim for algorithms where $\gamma_x$ decays exponentially in $k$, i.e., $\gamma_x \leq (g(q_x))^k$ for a polynomial $g$ with high probability. 

The focus on points where the single-run error is bounded (e.g., $q_x < 1/2$) can be justified. The following observation confirms that this condition is not a restrictive assumption, but rather a guaranteed property for the vast majority of the domain for any valid PAC learner.

\begin{restatable}{observation}{obsQxBound}
\label{obs:qx_bound}
Let $\mathcal{A}$ be an $(\epsilon, \delta)$-learner. Then, the single-run error $q_x$ satisfies $q_x < 1/2$ for at least a $1 - 2(\epsilon + \delta)$ fraction of the distribution $\mathcal{D}$.
\end{restatable}

\section{Sample Complexity Lower Bound}\label{sec:singleton}
In this section, we lower bound the sample complexity required to guarantee $(\gamma,\epsilon,\delta)$-individual monoculture robustness. Given that such algorithms must already satisfy standard $(\epsilon,\delta)$-learnability (which requires $n_{\mathcal{H}}(\epsilon,\delta)$ samples), we quantify the cost of robustness by lower bounding the necessary sample size multiplier, $t$, and ask:
\begin{center}
\textbf{What is the minimal $t$ required for a sample of size $t \cdot n_\mathcal{H}(\epsilon,\delta)$ to guarantee $(\gamma, \epsilon, \delta)$-individual monoculture robustness?}
\end{center}
We establish a worst-case lower bound on $t$, demonstrating that a logarithmic dependence on $1/\gamma$ is strictly necessary in this setting.
\begin{restatable}{theorem}{restateTheoremlowerBound}
\label{thm:lower_bound}
For every $d\ge 1$, every $k<\frac{1}{2\epsilon}$, and every
$\delta \leq \frac{\gamma}{4k}$, there exists a hypothesis class
$\mathcal{H}$ with $\mathrm{VC}(\mathcal{H})=d$, a distribution
$\mathcal{D}$ with $f^*\in \mathcal{H}$, and a point $x\in \mathcal{X}$  such that for any $(\epsilon,\delta)$-learning algorithm $\mathcal{A}$, the sample size multiplier $t$ required for a single random sample $S$ of size $t\cdot n_{\mathcal{H}}(\epsilon,\delta)$ to guarantee individual monoculture error
$
\Pr_{\mathcal{D},\mathcal{A}}
\left[e_k(\mathbf{h},x)=1\right]\le \gamma
$
is lower bounded by
$
t=\Omega\left(\log\frac{1}{\gamma}\right).
$
\end{restatable}

The theorem implies that the sample complexity for individual monoculture robustness scales as $\log(1/\gamma)\cdot n_\cH(\epsilon,\delta)$. 

We describe the proof using a singleton hypothesis class $\mathcal{H}=\{h_z \mid z\in\mathcal{X}\}$ where $h_z(x)=\1[x=z]$, defined over a finite domain with a uniform marginal distribution. 
To obtain arbitrary finite VC dimension $d\geq 1$, the full proof adds $d-1$ padding points that can be labeled arbitrarily and have zero probability under the marginal distribution.
The target function is $f^*=h_{x^*}$, where $x^*$ is selected uniformly at random from $\mathcal{X}$. We then analyze the probability that the target point $x^*$ is entirely missing from the training sample. The probability of this event decreases exponentially with the sample-size multiplier $t$. If $x^*$ is missing from the sample, the learner
sees only $0$ labels, and therefore has no information distinguishing $x^*$ from the other unseen
points. 

With high probability, each returned hypothesis has population loss at most $\epsilon$. On this good event, having loss at most $\epsilon$ with respect to the singleton target implies that each returned hypothesis can predict $1$  only on a small fraction of the domain; otherwise it would incur too many errors. Hence, the union of the points on which any of the $k$ hypotheses predicts $1$ cannot cover too many unseen points. Since the hidden target $x^*$ is uniform among the unseen points, we show that with probability at least $1/2$, none of the $k$ hypotheses predicts $1$ on $x^*$, so all $k$ hypotheses fail simultaneously. Thus, unless $t=\Omega(\log(1/\gamma))$, the individual monoculture error remains larger than $\gamma$.

A simple interpretation of the lower-bound construction is a rare qualified subpopulation of mass $p$. If the shared training sample contains no member of this subpopulation, all $k$ models may reject the same qualified individual because the data provide no evidence distinguishing that subpopulation. This event has probability $(1-p)^{tn}$. For a baseline sample size $n=\Theta(1/p)$, reducing it below $\gamma$ requires $t=\Omega(\log(1/\gamma))$. Thus, the construction captures a practical failure mode: models can have high overall accuracy while repeatedly failing on individuals from a subpopulation missing from their shared data.
\section{Monoculture Robustness via Replicability}\label{sec:replicability}
Consider a randomized learning algorithm $\mathcal{A}$ that, given a sample of size $n$, returns a single hypothesis. A key observation motivating this section is that monoculture errors stem from two distinct sources: the internal randomness of the algorithm and the randomness of the sample.

In general, when the single-run error satisfies $q_x<1$, there can be two extreme causes for  high individual monoculture error:

\textbf{1. Error driven by sample randomness ($\cD$).}
In this case, the failure depends primarily on the realization of the training set $S$. If $S$ is unrepresentative or ``bad'' for a specific input $x$, the algorithm might misclassify $x$ for most of  its internal randomness. Consequently, running the algorithm with fresh internal seeds will provide a small  benefit, and the individual monoculture error remains high. This type of error  drives the lower bound that appears in Section~\ref{sec:singleton}.

\textbf{2. Error driven by internal randomness.}
In this case, for a typical training sample $S$, failure at $x$ is not inherent to $S$: 
changing the internal random seed is likely to change the prediction, since many seeds lead to the correct label.
Therefore, running the learner $k$ times on the same sample with fresh random seeds can significantly reduce the joint
failure probability.


We capture the latter case via \textit{replicability}, which formalizes that for each fixed seed $r$,
the prediction of $\cA$ on $x$ is stable across most samples $S$. 
This lets us 
argue that on a good sample, a fresh randomness seed fails only with small probability. 
An algorithm is called \textit{$\rho$-replicable} if, by holding its internal randomness $r$ fixed and running it on two independent samples $S_1, S_2$ drawn from the same distribution $\cD$, it returns identical outputs (hypotheses) with probability at least $1-\rho$.

\begin{definition}[Replicability,~\cite{Impagliazzo22}]
\label{def:rep}
For $\rho\in[0,1/2)$, a randomized algorithm $\cA$ is $\rho$-replicable if for every distribution $\cD$,
\[
\Pr_{\substack{S_1,S_2\sim \cD^n,r}}\bigl[\cA(S_1; r) = \cA(S_2; r)\bigr]
\;\ge\;1 - \rho.
\]
\end{definition}
To formally address base learners that are replicable, we define the following \textit{Same-Sample Meta-Algorithm}: given a single dataset $S$, we execute the base algorithm $\cA$ on $S$ independently $k$ times (with fresh internal randomness) to generate a tuple of hypotheses $(h_1, \dots, h_k)$. 

Replicability alone is insufficient. It merely ensures that the failure probability is determined mostly by the internal seed. If the algorithm is biased towards the wrong label across most seeds, replicability will simply make it \textit{consistently} wrong. Therefore, to guarantee individual monoculture robustness via the meta-algorithm, we rely on  the base algorithm $\mathcal{A}$ to be sufficiently accurate on individuals. Specifically, we require the single-run error $q_x$, which depends on the algorithm $\mathcal{A}$, to be small (e.g., $q_x \le 1/2$).


While the notion of replicability is valuable on its own merit, it is a stringent requirement that is often impossible to satisfy (e.g., even for simple one-dimensional threshold functions, see~\citet{bun23rep}). A weaker notion called \textit{Pointwise Replicability}, recently introduced by \citet{hopkins2025approximate}, relaxes this condition. While replicability demands that the algorithm outputs the same hypothesis $h$ w.h.p., pointwise replicability only requires that the algorithm's predictions be consistent on every fixed individual $x$.

\begin{definition}[Pointwise Replicability,  \citet{hopkins2025approximate}]
\label{def:pointwise-rep}
For $\rho\in[0,1/2)$, an algorithm $\mathcal{A}$ is $\rho$-pointwise replicable if for every distribution $\cD$ and every $x\in\cX$, 
\[
\Pr_{S_1,S_2\sim \cD^n,r}\bigl[\cA(S_1; r)(x) = \cA(S_2; r)(x)\bigr] \ge 1 - \rho.
\]
\end{definition}
We are going to rely on this weaker notion for our results in this section for two reasons. First, replicability strictly implies pointwise replicability (the converse does not hold), so by showing that pointwise replicability implies individual monoculture robustness w.h.p. we also show that replicability implies individual monoculture robustness w.h.p.. 
Second, \citet{hopkins2025approximate} show that pointwise replicability is feasible for a much broader class of problems than replicability. Specifically, unlike replicability, \textit{any} hypothesis class with finite VC dimension admits a pointwise replicable learner (see discussion at the end of the section).

We now show that pointwise replicability is sufficient to guarantee monoculture robustness. To facilitate the analysis, let $R$ denote the set of the algorithm's internal random seeds, and let $\mathcal{S}$ denote the (possibly infinite) multiset of samples representing the distribution $\mathcal{D}^n$. I.e., $\mathcal{S}$ represents the sample space weighted by the probability mass of $\mathcal{D}^n$.

\subsection{The Monomatrix Argument}



We adopt a matrix-based perspective (Definition~\ref{def:monomatrix}) we call monomatrix to disentangle the two sources of randomness. Our argument proceeds by characterizing ``stability'' via the rows and ``representativeness'' via the columns:
\begin{enumerate}
\item \textbf{Defining Canonical Behavior (Row Majority):} First, we examine the rows (internal seeds) of the monomatrix. Pointwise replicability implies that for any fixed seed $r$, the algorithm's output on each point in the domain is stable across most samples. This allows us to define a \textit{row majority} value $v_r$—the ``canonical'' label that seed $r$ outputs most of the time. In Lemma~\ref{lem:few-wrong-rows}, we show that if the algorithm is accurate on $x$ (small $q_x$), then most seeds usually vote for the correct label (i.e., $v_r = f^*(x)$).
\item \textbf{Identifying Representative Data (Good Samples):} Next, we examine the columns (samples). We need to ensure that the specific sample $S$ we draw preserves the canonical behavior of the seeds. We define a \textit{good sample} (formalized as not $\tau$-bad) as one where the algorithm outputs the canonical label $v_r$ for the vast majority of seeds $r$. Lemma~\ref{lem:column-bound} establishes that such samples are highly probable, meaning strictly ``bad'' samples (that trigger widespread deviation from the canonical outputs) are rare.

\item \textbf{Synthesis (Monoculture Robustness):} Finally, Theorem~\ref{thm:mono_error_non_bad} combines the two previous steps. Conditioned on drawing a good sample (which is likely), the learner's error is driven purely by the small fraction of seeds that have the \textit{wrong} canonical majority. Since these seeds are rare, the probability that $k$ independent runs all hit these bad seeds decays exponentially.
\end{enumerate}

We begin. To systematically analyze the interaction between internal randomness and sample randomness for a fixed input $x$, we organize the algorithm's correctness on an individual $x$ into a matrix structure defined over these two sources of randomness. See illustration in Table~\ref{table:monomatrix}.

\begin{definition}[Monomatrix]
\label{def:monomatrix}
Fix $x\in \mathcal{X}$. The \textit{monomatrix} of $x$, denoted $M_x$, is the $\{0,1\}$-valued array indexed by $r\in R$ and $S\in \cS$ defined by
\[
[M_x]_{r,S} \;=\; \1\!\big[\mathcal{A}(S; r)(x) = f^*(x)\big].
\]
We note that while we visualize this as a matrix with columns in $\cS$, which can technically be infinite, our analysis treats $S$ as a random variable distributed according to $\mathcal{D}^n$.
\end{definition}

Pointwise replicability implies that for any fixed internal seed $r$, the algorithm's prediction on $x$ is stable across most samples. 
In the monomatrix view, this corresponds to a row-majority label property: most entries in row $r$ share the same value.

\begin{definition}[Row-majority and $\tau$-bad samples]
\label{def:tau-bad}
Fix a point $x$. For each internal seed $r \in R$, let $v_r \in \{0,1\}$ be the value that is most likely to be output by the algorithm over the choice of sample $S \sim \mathcal{D}^n$:
\[
v_r \;=\; \mathop{\mathrm{argmax}}_{y \in \{0,1\}} \Pr_{S \sim \mathcal{D}^n} [\mathcal{A}(S;r)(x) = y].
\]
For a sample $S$, define the deviation score $\Delta(S)$ as the fraction of seeds that disagree with their row majority:
\[
\Delta(S) \;=\; \frac{1}{|R|}\sum_{r \in R} \1\big[\mathcal{A}(S;r)(x) \neq v_r\big].
\]
We say that a sample $S$ is \textit{$\tau$-bad} if $\Delta(S) > \tau$.
\end{definition}

\begin{table}[ht]
\begin{center}
\setlength{\extrarowheight}{4pt}
\begin{tabular}{c|c|c|c|c|}
    \multicolumn{1}{c}{} & \multicolumn{4}{c}{$\cS$} \\
    \cline{2-5}
    \multirow{3}{*}{R} & & $\vdots$ & & \\
    \cline{2-5}
    & \multicolumn{1}{c}{\cellcolor{blue!20}{$v_r=\1[\cA(s,r)(x) = f^*(x)]$}}
    & \multicolumn{1}{c}{\cellcolor{blue!20}{$\hdots$}}
    & \multicolumn{1}{c|}{\cellcolor{blue!20}}
    & $\le \rho$ \\
    \cline{2-5}
    & & $\vdots$ & & \\
    \cline{2-5}
\end{tabular}
\end{center}
\caption{Illustration of a monomatrix for a replicable algorithm, $\cA$. The solid entries represent rows where the majority prediction $v_r$ is the same across samples.}
\label{table:monomatrix}
\end{table}

The following lemma utilizes Markov's inequality to translate the pointwise replicability parameter into a bound on the fraction of bad columns in the monomatrix.

\begin{restatable}
{lemma}{restateColumnBound}
\label{lem:column-bound}
Suppose $\cA$ is $\rho$-pointwise replicable. Then for every $\tau \in (0,1)$,
$$
\Pr_{S \sim \mathcal{D}^n} [S \text{ is } \tau\text{-bad}] \;\le\; \frac{\rho}{\tau}.
$$
\end{restatable}

While Lemma \ref{lem:column-bound} controls how often the algorithm deviates from its majority behavior, we must also quantify the error rate of the majority vote itself.

\begin{definition}[Row-majority error at $x$]
Define the row-majority error at $x$ by
\[
m \;:=\; \frac{1}{|R|}\,\#\bigl\{r \in R : v_r \neq f^*(x) \bigr\}.
\]
\end{definition}

By union bound, the single-run error on any sample $S$ is at most the sum of the majority error rate $m$ and the deviation rate $\Delta(S)$. Consequently, for any ``good'' sample (where $\Delta(S) \le \tau$), we obtain the following guarantee:

%
\begin{restatable}{corollary}{restateCorMonoError}
\label{cor:mono-error-on-fixed-sample}
If a sample $S$ is not $\tau$-bad, then for any $k \in \mathbb{N}$, running $\cA(S;\cdot)$ independently $k$ times yields
$$
\Pr\big[e_k(\textbf h,x)|\, S\big] \;\le\; \big(m+\tau\big)^{k}.
$$
\end{restatable}

To utilize Corollary \ref{cor:mono-error-on-fixed-sample}, we must bound $m$. The following lemma shows that if   $q_x$, the algorithm's  baseline misclassification error  on $x$, is sufficiently small, then the fraction of internal seeds yielding the wrong majority must also be small.

\begin{restatable}
{lemma}{restateFewWrongRows}
\label{lem:few-wrong-rows}
Let $u_r:=\Pr_{S}[\cA(S;r)(x)\neq f^*(x)]$ be the fraction of wrong classification values in a row $r$.
If $q_x=\E_{r}[u_r]<\tfrac12$, then 
$$
\Pr_{r}[v_r \neq f^*(x)] \le 2q_x.
$$
\end{restatable}

We now synthesize Lemmas~\ref{lem:column-bound} and~\ref{lem:few-wrong-rows}, and Corollary~\ref{cor:mono-error-on-fixed-sample} 
 into the main theorem of this section.

\begin{theorem}
\label{thm:mono_error_non_bad}
Assume $q_x<1/2$. Let $\mathcal{A}$ be a $\rho$-pointwise replicable $(\epsilon,\delta)$-learner. Let $\mathcal{M}$ be the same-sample meta-algorithm that, given a single sample $S \sim \mathcal{D}^n$, runs $\mathcal{A}$ independently $k$ times with fresh internal randomness to return an ensemble $h = (h_1, \dots, h_k)$. Fix $\tau \in (0, 1)$.

If a sample $S$ is not $\tau$-bad, then the conditional individual monoculture error is bounded by:
\[
\Pr[e_k(h, x) = 1 \mid S] \leq (2q_x + \tau)^k.
\]
Consequently, with probability at least $1 - \rho/\tau$ over the choice of $S \sim \mathcal{D}^n$, this bound holds. In particular, for $\tau = \sqrt{\rho}$, with probability at least $1 - \sqrt{\rho}$:
\[
\Pr[e_k(h, x) = 1 \mid S] \leq (2q_x + \sqrt{\rho})^k.
\]
\end{theorem}

\begin{proof}
Consider a sample $S$ that is not $\tau$-bad. By Corollary~\ref{cor:mono-error-on-fixed-sample}, the failure probability is bounded by $(m+\tau)^k$.
Using Lemma~\ref{lem:few-wrong-rows}, if $q_x \le \tfrac12$, then $m \le 2q_x$, yielding the bound $(2q_x+\tau)^k$.
Finally, Lemma~\ref{lem:column-bound} guarantees that $S$ is not $\tau$-bad with probability at least $1-\rho/\tau$.
\end{proof}
This means that if we have a (pointwise) replicable algorithm which is sufficiently accurate on $x$, w.h.p. over the draw of the training set $S$, the individual monoculture error on $x$ decays exponentially with the ensemble size $k$. 
\begin{remark}
While the above theorem is stated pointwise,  
Observation~2.2 implies that the condition 
$q_x<1/2$ holds for at least a $1-2(\epsilon+\delta)$ mass of points under $\mathcal D_X$. Therefore, for each point $x$ in this set, \cref{thm:mono_error_non_bad} gives a high-probability guarantee over the draw of $S$.
\end{remark}
Next, we discuss the efficiency of achieving individual monoculture robustness by designing a  pointwise replicable algorithm. We first state a result from~\cite{hopkins2025approximate}.
\begin{theorem}
\label{thm:hopkins}
Let $\mathcal H$ be a hypothesis class with VC dimension $d$.
For any $\rho,\epsilon,\delta \in (0,1)$ there exists
a $\rho$-pointwise replicable learner that, with probability at least 
$1-\delta$, outputs a hypothesis with error at most $\inf_{h\in \cH}\cL_D(h)+\epsilon$. 
The sample complexity is 
$O\!\left(\frac{d \log(1/\delta)}{\rho^{3}\epsilon}\right)$ in the realizable setting, and $O\!\left(\frac{d \log(1/\delta)}{\rho^{4}\epsilon^2}\right)$ in the agnostic setting.
In addition, there exists a constant $c\in \mathbb{R}$ such that any $\rho$-pointwise replicable $(\epsilon, c)$-learner requires $\Omega \left( \frac{d}{\rho^2 \epsilon \sqrt{\log(1/\rho)}} \right)$ samples in the realizable setting and $\Omega \left( \frac{d}{\rho^2 \epsilon^2 \sqrt{\log(1/\rho)}} \right)$ in the agnostic setting .
\end{theorem}

By combining the robustness guarantee of Theorem~\ref{thm:mono_error_non_bad} with the above, we get a bound over the sample complexity required to ensure individual monoculture robustness for any hypothesis class.

Monoculture robustness is a weaker notion than pointwise replicability. Consequently, (pointwise) replicable algorithms provide individual monoculture robustness ``for free''. This reveals a counter-intuitive connection: enforcing consistency across samples guarantees diversity across internal seeds for a fixed sample. However, unless replicability is explicitly required,
a sample partitioning meta algorithm that runs on a sample of size $kn$ is practically superior
. It avoids the heavy sample complexity of replicability (scaling with $1/\rho^2$ ), making it the efficient choice in the typical regime where $k \ll 1/\rho^2$.

\section{Discussion} 
In this paper, we introduced a learning framework to rigorously quantify algorithmic monoculture in learning theory. By formalizing the notion of conjunctive error within the PAC setting, we disentangled the two primary sources of correlated failure: sampling and internal algorithmic randomness. This reveals a fundamental dichotomy in addressing monoculture. 

When errors are driven by the specific realization of the training data (e.g., the absence of a subpopulation in the sample), our lower bound (Theorem~\ref{thm:lower_bound}) establishes that 
 learning cannot, in general, manufacture independence from missing information.
In this regime, the ``shared blind spot'' is information-theoretic; to drive the individual monoculture error down in this model, a logarithmic multiplier in sample size is unavoidable. This suggests that for hard inputs, organizational redundancy (running $k$ models) provides false security unless backed by genuinely independent data streams.

Conversely, when errors stem from internal randomness, we uncovered a surprising utility for Replicability. While typically studied as a tool for scientific reproducibility or privacy, we showed that replicability essentially ``stabilizes'' the learner’s behavior on good samples, thereby ensuring that fresh random seeds yield independent trajectories of success. This reframes replicability not just as a constraint for stability, but as a sufficient condition to control conjunctive errors. 
Our monomatrix perspective gives a useful viewpoint on replicability, not explicitly present in the literature to the best of our knowledge: it separates sample randomness from internal randomness and interprets replicability as a row-stability property, which may also be useful for studying how stability interacts with randomness, diversity, and failure correlation more broadly.

Finally, our interpretation of the~\cite{Aden-AliHLZ24} sample-splitting lemma under the expected-monoculture-error objective in Appendix~\ref{sec:expected} offers a practical prescription. 
In settings where the data budget is fixed,  training two weaker hypotheses on independent halves of the data often yields a lower joint failure rate than training a single duplicated hypothesis on the full dataset. This challenges the prevailing dogma that maximizing individual model accuracy is the sole path to system reliability. It suggests that in high-stakes environments where joint failure is catastrophic (e.g., loan approvals), deliberately inducing independence via data partitioning may be superior to maximizing the marginal accuracy of a single monoculture.

\textbf{Future Work}. 
Future work includes developing ``monoculture-aware'' training objectives that directly penalize correlated errors without requiring strong replicability constraints, extending data-partitioning guarantees to the agnostic case, and studying settings beyond binary classification, such as multiclass classification, regression, ranking, and generative tasks. Another important direction is to analyze deployment settings where independently trained models share pretrained components (see Appendix~\ref{app:shared-pretrained-components}), training data, or other parts of the training pipeline~\citep{Bommasani22Homogenization}.

\section*{Acknowledgments}

Lee Cohen and Omer Reingold are  supported by the Simons Foundation Collaboration on the Theory of Algorithmic Fairness, and the Simons Foundation investigators award 17351.





\nocite{langley00}

  \bibliography{ref}
  \bibliographystyle{apalike}

\newpage
\appendix


\section{Minimizing Expected Monoculture Error via Sample Splitting}\label{sec:expected}
Consider the realizable, proper learning setting. When minimizing the expected monoculture error, a natural question arises regarding the optimal use of a fixed data budget when training an ERM. Given a sample of size $2n$, is it better to train a single, highly accurate hypothesis on all the data and use it twice, or to partition the data into two independent sets of size $O(n)$ and produce separate hypotheses? 
In this section, we interpret this question through an existing sample-partitioning result of~\cite{Aden-AliHLZ24}. Their Lemma 5 implies that, in the realizable proper learning setting, splitting the sample into two independent halves and running ERM on each yields expected monoculture error $O(d/n)$, effectively gaining a logarithmic factor over the standard expected error bound for a single ERM trained on the full dataset.

For completeness, we start by recalling the expected error bound for ERM by~\cite{HAUSSLER1994248}.
\begin{fact}[ERM expected error]\label{fact:ERM-expected-err}
For realizable classes of VC dimension $d$, ERM satisfies
\[
\sup_{\cD}\mathbb{E}_{S\sim \cD^n}[\cL_D(\mathcal A(S))]=O\Big(\tfrac{d}{n}\log\tfrac{n}{d}\Big).
\]
Moreover, there exist $\mathcal H,\cD$ where the lower bound $\mathbb{E}_{S\sim \cD^n}[\cL_D(\mathcal A(S))]=\Omega\Big(\tfrac{d}{n}\log\tfrac{n}{d}\Big)$ 
holds.
\end{fact}

Combining the expected-monoculture-error viewpoint with~\cite{Aden-AliHLZ24},  one obtains a logarithmic-factor improvement relative to the standard single-ERM expected-error benchmark. 
To maintain consistency with the definition of $q_x$ (which assumes a training set of size $n$), we assume a total data budget of $2n$. While this scales the sample size by a constant factor, it does not alter the asymptotic order of the convergence bounds.

\begin{restatable}
{fact}{restateExpectedMono}[\cite{Aden-AliHLZ24}, Lemma 5, restated for conjunctive error]
\label{thm:expected-mono}
Let $\mathcal A$ be a sample partitioning meta algorithm that uses a fixed ERM algorithm $A$ on a realizable class $\mathcal H$ with VC dimension $d$. Given a sample $S$ of $2n$ i.i.d. samples, algorithm $\cA$ splits $S$ uniformly at random into two disjoint halves $S_1,S_2$ of size $n$ and returns $h_1= A(S_1)$, $h_2= A(S_2)$. Then,
\[
\mathbb{E}_{x\sim D}\big[e_2((h_1,h_2)^T,x)\big]
=O\Big(\tfrac{d}{n}\Big).
\]
\end{restatable}

We remark that they take the majority of $3$ such hypotheses (which is no longer proper), and guarantee an error rate of $O(d/n)$ using this lemma.

This interpretation of~\cite{Aden-AliHLZ24} connects our expected monoculture-error objective to boosting and optimal learning literature (e.g., \citet{Schapire90, Freund95boosting, hanneke2016refined}). In those works, aggregation methods are used to reduce the marginal error of a final improper predictor. 
Here, we use the same underlying independence phenomenon differently: rather than aggregating the hypotheses into one predictor, we keep the (ERM proper) hypotheses separate and evaluate their joint failure probability. 

Thus, two proper ERMs trained on disjoint halves may have joint failure $O(d/n)$, even though duplicating a single ERM would inherit the larger standard expected-error benchmark.

\paragraph{Agnostic learning and label noise.}
For two learners trained on independent sample halves, their expected conjunctive error is $\mathbb{E}_x[q_x^2]$, whereas standard agnostic PAC guarantees control only the marginal error $\mathbb{E}_x[q_x]$. From Jensen's inequality,
\[
\bigl(\mathbb{E}_x[q_x]\bigr)^2
\leq \mathbb{E}_x[q_x^2]
\leq \mathbb{E}_x[q_x].
\]
Misspecification errors may be shared across learners or may occur on different inputs, so conjunctive error may be close to either endpoint. A sharper agnostic rate therefore requires assumptions controlling the overlap of approximation errors, not only marginal excess risk. Our current framework assumes a fixed target $f^*$. Explicitly modeling stochastic label noise or irreducible Bayes error would require an extension because the shared label randomness can itself induce correlated failure.

\section{Additional Related Works}\label{app:related}

\textbf{Model multiplicity and predictive multiplicity.}
Our work is related to the existence and characterization of near-optimal models \citep{Breiman01,Fisher19,Paes23,Semenova22}, and to predictive multiplicity, underspecification, and empirical arbitrariness \citep{Black22Multiplicity,DAmour22,Gomez24,Watson-Daniels23}.

\citet{Paes23} study the number of samples needed to statistically distinguish the performance of models in a Rashomon set, while \citet{Semenova22} study when a large Rashomon set contains (accurate) simpler models. These results concern properties of a set of near-optimal models. Our question begins after such a set may exist: can a learner with a single dataset produce $k$ accurate hypotheses whose errors are not jointly concentrated on the same individuals?

Thus, the two perspectives are complementary rather than opposites. A large Rashomon set provides potential diversity that a monoculture-robust algorithm might exploit. However, multiplicity alone does not imply monoculture robustness. Many structurally distinct or equally accurate models may still misclassify the same point $x$, resulting in high conjunctive error. Conversely, low monoculture error does not require models to differ substantially everywhere. The hypotheses may agree on almost the entire domain and differ only in ways that prevent their errors from overlapping.

Moreover, existence results for the Rashomon set do not by themselves provide a procedure for selecting $k$ models with complementary errors. Our framework adds these algorithmic and learning-theoretic requirements.

\textbf{Algorithmic Fairness in Composition and Sequential Systems.} Our work is also closely related to and motivated by the rigorous study of fairness. \citet{Dwork12Fairness} introduced the notion of individual fairness, positing that fairness requires treating similar individuals similarly according to a task-specific metric. \citet{Yona18Metric} subsequently adapted this to the statistical learning setting, providing sample complexity bounds for learning hypotheses that are ``probably approximately'' metric-fair on the population. Algorithmic fairness has also been studied in the context of multiple classifiers. 

\citet{Dwork19Composition} formally analyzed fairness under composition, identifying that classifiers which are individually fair in isolation do not necessarily satisfy fairness guarantees when composed into a larger system. This line of inquiry has been extended to sequential settings and other fairness notions; for instance, \citet{Blum22Multi} showed that standard fairness constraints applied to individual stages of a hiring pipeline can paradoxically reduce the overall diversity of the selected cohort. Similarly, \citet{cohen20Screening} further demonstrated that when individual people come from groups with different noise levels, administering the screening policy to all candidates leads to disparate outcomes, necessitating dynamic policies that test noisier groups more often to achieve true equality. Relatedly,~\cite{Dai21FairnessPartialCompliance} study fair machine learning under partial compliance, showing that when only some employers adopt fairness-conscious policies, strategic applicant behavior can lead to sublinear fairness gains and segregation across institutions.

\citet{Liu19Delayed} and \citet{Kannan19Downstream} demonstrate that single-step fairness constraints can fail to promote long-term equity when decisions are repeated or interactive. More recently, \citet{gradwohl2025fairness} studied competitive environments, showing that even when fair firms are competing, the resulting ecosystem equilibrium may still be unfair, and that unilaterally improving one classifier's fairness can degrade the ecosystem's overall equity. 
Unlike these works, which analyze the composition of existing classifiers, our paper focuses on the learning of an ensemble that is explicitly robust to correlated errors. 

\textbf{Replicability in Learning.} Replicability was formally introduced into learning theory by~\citet{Impagliazzo22} as an algorithmic property to address the reproducibility crisis in science; it requires that a randomized algorithm produces the exact same output with high probability when run on two independent samples from the same distribution. This notion was subsequently shown to be intimately connected to differential privacy and adaptive data analysis: \citet{bun23rep} proved that replicability is equivalent to approximate differential privacy in the finite sample regime, establishing that "stability is stable" and implying strong generalization bounds. To characterize the limits of this stability, \citet{Kalavasis23} introduced the notion of Total Variation (TV) indistinguishability, providing information-theoretic equivalences between replicability and other stability concepts. While strict replicability can be restrictive—imposing polynomial sample complexity overheads or impossibility results for basic tasks—recent work by~\citet{hopkins2025approximate} explores relaxations such as approximate and pointwise replicability. These weaker notions retain the spirit of reproducible science while enabling sample-efficient learning for any class with finite VC dimension.

\textbf{Self-consistency}
Our problem is related to the notion of self-consistency~\cite{Cooper23Self-Consistency}. For a fixed input $x$, self-consistency is defined as $SC(x) = \Pr_{S,S'}[A(S)(x) = A(S')(x)]$, where the two models are produced by the same learning process using different training samples. This  notion is label-independent and measures stability or lack of arbitrariness. In contrast, our individual monoculture error is $\Pr_{S,\mathcal{A}}[e_k(h,x) = 1]$. It is label-dependent and measures the joint failure of $k$
 individually accurate predictors.
 
The objectives are aligned when consistency reflects correctness. If all models agree on the correct label, self-consistency is maximal and monoculture error is zero. However, agreement alone is insufficient. Models may consistently predict the wrong label, giving maximal self-consistency and, in our terminology, maximal monoculture failure. Conversely, disagreement can reduce monoculture error when it ensures that at least one model is correct, but diversity for its own sake may merely introduce incorrect or arbitrary predictions. Thus, monoculture robustness does not seek disagreement itself. It seeks diversity specifically in the models’ errors while maintaining their individual PAC guarantees.

There is also a direct technical connection. Under~\cite{Cooper23Self-Consistency} setup, the varying source is the training sample (simulated via bootstrap replicates). This makes their self-consistency condition closely resemble the pointwise agreement across samples captured by pointwise replicability. The key technical distinction is how each handles internal randomness: pointwise replicability achieves this cross-sample stability by explicitly fixing the algorithm's internal randomness seed, whereas self-consistency does not condition on a shared internal seed. Our main theorem leverages this exact distinction: we use seed-conditioned stability across samples, together with individual accuracy, to prove that running the algorithm multiple times with fresh internal seeds on a fixed shared sample yields an individual monoculture error that decreases exponentially with $k$
.

Therefore, stability across samples (for a fixed seed) can coexist with useful diversity across internal seeds (for a fixed sample). The two goals consequently address complementary deployment concerns. \cite{Cooper23Self-Consistency} seek to avoid arbitrary individual decisions, using ensembling and abstention when agreement is insufficient. We seek to prevent several decision-makers from sharing the same harmful error. Combining high self-consistency across samples with low joint failure across independently randomized predictors is an interesting direction for future work. 

\section{Omitted Proofs}\label{sec:proofs}
\obsQxBound*
\begin{proof}
Recall that $q_x = \Pr_{S \sim \mathcal{D}^n}[\mathcal{A}(S)(x) \neq f^*(x)]$ is the probability that the algorithm errs on $x$ over the randomness of the sample. By linearity of expectation, the expected value of $q_x$ over the domain is equivalent to the expected generalization error of the algorithm:
\[
    \mathbb{E}_{x \sim \mathcal{D}}[q_x] = \mathbb{E}_{x \sim \mathcal{D}}\mathbb{E}_{S \sim \mathcal{D}^n}[\mathbb{I}(\mathcal{A}(S)(x) \neq f^*(x))] = \mathbb{E}_{S \sim \mathcal{D}^n}[\mathcal{L}_{\mathcal{D}}(\mathcal{A}(S))].
\]
Since $\mathcal{A}$ is an $(\epsilon, \delta)$-learner, we know that $\Pr_S[\mathcal{L}_{\mathcal{D}}(\mathcal{A}(S)) \leq \epsilon] \geq 1 - \delta$. Bounding the loss by $1$ in the failure event, we have:
\[
    \mathbb{E}_{S}[\mathcal{L}_{\mathcal{D}}(\mathcal{A}(S))] \leq \epsilon(1-\delta) + 1 \cdot \delta \leq \epsilon + \delta.
\]
Applying Markov's inequality to $q_x$, we obtain:
\[
    \Pr_{x \sim \mathcal{D}}[q_x \geq 1/2] \leq \frac{\mathbb{E}_x[q_x]}{1/2} \leq 2(\epsilon + \delta).
\]
Thus, $q_x < 1/2$ holds with probability at least $1 - 2(\epsilon + \delta)$ over the choice of $x$.
\end{proof}

\subsection{Omitted Proofs from Section~\ref{sec:singleton}}
\restateTheoremlowerBound*
\begin{proof}

Fix $d\ge 1$. Consider a domain of the form
\[
\mathcal{X}=\mathcal{X}_{\mathrm{hard}}\sqcup \mathcal{X}_{\mathrm{pad}},
\]
where
\[
\mathcal{X}_{\mathrm{hard}}=\{1,\dots,M\}
\]
and
\[
\mathcal{X}_{\mathrm{pad}}=\{p_1,\dots,p_{d-1}\}.
\]
The marginal distribution is uniform on $\mathcal{X}_{\mathrm{hard}}$ and assigns probability $0$
to $\mathcal{X}_{\mathrm{pad}}$.

Define the hypothesis class $\mathcal{H}$ as follows. For every
$z\in \mathcal{X}_{\mathrm{hard}}$ and every labeling
$b\in\{0,1\}^{d-1}$ of the padding points, let $h_{z,b}$ be the hypothesis satisfying
\[
h_{z,b}(x)=\mathbb{I}[x=z] \qquad \text{for } x\in \mathcal{X}_{\mathrm{hard}},
\]
and
\[
h_{z,b}(p_j)=b_j \qquad \text{for } j=1,\dots,d-1.
\]
Thus, on $\mathcal{X}_{\mathrm{hard}}$, the class behaves like the singleton class, while the
padding points can be labeled arbitrarily.

We have $\mathrm{VC}(\mathcal{H})=d$. Indeed, for any  $p'\in\cX_{\mathrm{hard}}$ the set
$\{p',p_1,\dots,p_{d-1}\}$ is shattered: the padding points can be labeled arbitrarily, and the
point $p'$ can be labeled either by choosing $z=p'$ or by choosing some $z\neq p'$. On the other hand,
no set of size $d+1$ can be shattered, since such a set must contain at least two points from
$\mathcal{X}_{\mathrm{hard}}$, and the class cannot label two distinct hard-domain points by $1$
simultaneously.

Let the target function be $f^*=h_{x^*,0}$, where $x^*$ is selected uniformly at random from
$\mathcal{X}_{\mathrm{hard}}$, and $0$ denotes the all-zero labeling of the padding points.

Let $n=n_{\mathcal{H}}(\epsilon,\delta)$ be the sufficient sample size for a standard
$(\epsilon,\delta)$-learner for this class $\mathcal{H}$. Let
\[
T := \left\lfloor \alpha \log \frac{1}{\gamma} \right\rfloor
\]
for a sufficiently small universal constant $\alpha>0$, and set
\[
M=\left\lceil \frac{4Tn+8k}{1-2k\epsilon}\right\rceil .
\]
Since $k<1/(2\epsilon)$, the domain size is well-defined.

We prove the lower bound in expectation over the random choice of $x^*$ from
$\mathcal{X}_{\mathrm{hard}}$. It then implies that there exists a fixed
$x^*\in\mathcal{X}_{\mathrm{hard}}$ for which the same lower bound holds.

Consider a single random sample $S$ of size $N=tn$, where $t\le T$. Let $E$ be the event that the
target point $x^*$ is not present in $S$. Since the distribution is supported on
$\mathcal{X}_{\mathrm{hard}}$, the probability of this event is
\[
\Pr[E]=\Pr[x^*\notin S]=\left(1-\frac{1}{M}\right)^{tn}.
\]
Since $M\ge 2$, define
\[
c=\left(1-\frac{1}{M}\right)^n\in(0,1),
\]
so $\Pr[E]=c^t$.

By our choice of $M$, we have $M\ge 4Tn$, and therefore, since $1\le t\le T$,
\[
\frac{n}{M}\le \frac{tn}{M}\le \frac{1}{4}.
\]
By Bernoulli's inequality,
\begin{align}\label{eq:c}
c=\left(1-\frac{1}{M}\right)^n \ge 1-\frac{n}{M}\ge \frac{3}{4}.
\end{align}

Conditioned on the event $E$, the sample $S$ contains only $0$-labeled examples. The algorithm
$\mathcal{A}$ returns an ensemble of $k$ hypotheses $\{h_1,\dots,h_k\}$. The conjunctive error
event occurs if all hypotheses fail on $x^*$. Since $f^*(x^*)=1$, failure means
$h_i(x^*)=0$ for all $i=1,\dots,k$.

Let $G_i$ be the event that the $i$-th returned hypothesis has loss at most $\epsilon$, and let
$G=\cap_i G_i$. Since $\mathcal{A}$ is an $(\epsilon,\delta)$-learner, by a union bound,
\[
\Pr[G]\ge 1-k\delta.
\]

Let
\[
U=\bigcup_{i=1}^k \{x\in\mathcal{X}_{\mathrm{hard}}: h_i(x)=1\}
\]
be the set of hard-domain points predicted as $1$ by at least one hypothesis in the ensemble. We now
bound the probability that the ensemble covers the target point, namely the event $x^*\in U$.

The following observation is where we use the singleton construction on
$\mathcal{X}_{\mathrm{hard}}$. This argument applies also to improper hypotheses. Indeed, for any
 hypothesis $h$, if $L_D(h)\le \epsilon$ with respect to the singleton target $h_{x^*,0}$, then
\[
|\{x\in\mathcal{X}_{\mathrm{hard}}:h(x)=1\}|\le \epsilon M+1.
\]
To see this, if $h(x^*)=1$, then every point in
$\{x\in\mathcal{X}_{\mathrm{hard}}:h(x)=1\}\setminus\{x^*\}$ is a false positive, so
\[
|\{x\in\mathcal{X}_{\mathrm{hard}}:h(x)=1\}|-1\le \epsilon M.
\]
If $h(x^*)=0$, then all points in
$\{x\in\mathcal{X}_{\mathrm{hard}}:h(x)=1\}$ are false positives, and the bound is $\epsilon M$.
Therefore, on the event $G$,
\[
|U|\le \sum_{i=1}^k |\{x\in\mathcal{X}_{\mathrm{hard}}:h_i(x)=1\}|
\le k(\epsilon M+1).
\]

Now condition on the event $E$, on the realized unlabeled sample points, and on the internal
randomness of the learner. Since $x^*$ was chosen uniformly from
$\mathcal{X}_{\mathrm{hard}}$ and does not appear in $S$, its conditional distribution is uniform
over the unseen set $\mathcal{X}_{\mathrm{hard}}\setminus S$. Moreover, on the event $E$, the learner
only observes $0$ labels, so its output is determined by $S$ and its internal randomness, but not by
the identity of $x^*$ except through the fact that $x^*\notin S$.

We need to bound the probability of the event $G\cap\{x^*\in U\}$. For any fixed realization of
$S$ and of the learner's internal randomness, consider the set of unseen points $x$ for which both
$G$ would hold with target $h_{x,0}$ and $x\in U$. For every such point, the preceding observation
implies
\[
|U|\le k(\epsilon M+1).
\]
Therefore,
\[
\Pr[G\cap\{x^*\in U\}\mid E]
\le
\frac{k(\epsilon M+1)}{M-N}.
\]
Since $N=tn\le Tn$, our choice of $M$ gives
\[
M(1-2k\epsilon) > 2Tn+2k \ge N+2k.
\]
Equivalently,
\[
2k(\epsilon M+1)<M-N.
\]
Thus,
\[
\Pr[G\cap\{x^*\in U\}\mid E]
\le
\frac{k(\epsilon M+1)}{M-N}
<\frac{1}{2}.
\]

Therefore,
\[
\Pr[e_k(\mathbf{h},x^*)]
\ge
\Pr[E]-\Pr[G^c]-\Pr[E\cap G\cap \{x^*\in U\}]
\ge
c^t-k\delta-\frac{1}{2}c^t
=
\frac{1}{2}c^t-k\delta.
\]

Thus, to satisfy the robustness condition $\Pr[e_k(\mathbf{h},x^*)]\le \gamma$, we must have
\[
\frac{1}{2}c^t-k\delta\le \gamma,
\]
and hence
\[
c^t\le 2(\gamma+k\delta).
\]
Using the assumption $\delta\le \gamma/(4k)$, we get
\[
c^t\le \frac{5}{2}\gamma.
\]
Taking logarithms gives
\[
t\log\frac{1}{c}\ge \log\frac{2}{5\gamma}.
\]
By \cref{eq:c}, we have $c\ge 3/4$, and therefore
\[
\log\frac{1}{c}\le \log\frac{4}{3}.
\]
Consequently,
\[
t\ge \frac{\log(2/(5\gamma))}{\log(4/3)}
=
\Omega\left(\log\frac{1}{\gamma}\right).
\]


Thus, for every learner that uses sample-size multiplier
$t<\alpha\log(1/\gamma)$, for a sufficiently small universal constant $\alpha>0$, there exists a
target point $x^*\in\mathcal{X}_{\mathrm{hard}}\subseteq \cX$ such that the individual monoculture error at
$x^*$ is larger than $\gamma$. Equivalently, the sample-size multiplier required to guarantee
individual monoculture error at most $\gamma$ is
\[
t=\Omega\left(\log\frac{1}{\gamma}\right).
\]
\end{proof}
\subsection{Omitted Proofs from Section~\ref{sec:replicability}}

\restateColumnBound*
\begin{proof}

Fix a row (seed) $r \in R$. Recall that
\[
v_r =\arg\max_{y\in\{0,1\}} \Pr_{S\sim \mathcal D^n}\!\big[\mathcal A(S;r)(x)=y\big].
\]
Therefore, for this fixed $r$,
\begin{align*}    
\Pr_{S\sim \mathcal D^n}\!\big[\mathcal A(S;r)(x)\neq v_r\big]
&= 1-\max_{y\in\{0,1\}} \Pr_{S\sim \mathcal D^n}\!\big[\mathcal A(S;r)(x)=y\big]\\
&\le 1-\Pr_{S\sim \mathcal D^n}\!\big[\mathcal A(S;r)(x)=y_{x,r}\big]\\
&= \Pr_{S\sim \mathcal D^n}\!\big[\mathcal A(S;r)(x)\neq y_{x,r}\big],
\end{align*}
where $y_{x,r}=\text{argmax}_{y\in\{0,1\}} \Pr_{S\sim \mathcal D^n}\!\big[\mathcal A(S;r)(x)=y\big]$ is the pointwise-replicable label guaranteed by Definition~\ref{def:pointwise-rep}.

Taking expectation  over $r\sim R$ w.r.t. the seed distribution and using linearity of expectation,
\begin{align*}    
\mathbb E_{S\sim \mathcal D^n}[\Delta(S)]
&= \mathbb E_{r}\Pr_{S\sim \mathcal D^n}\!\big[\mathcal A(S;r)(x)\neq v_r\big]\\
&\le \mathbb E_{r}\Pr_{S\sim \mathcal D^n}\!\big[\mathcal A(S;r)(x)\neq y_{x,r}\big]\\
&= \Pr_{S\sim \mathcal D^n,\ r}\!\big[\mathcal A(S;r)(x)\neq y_{x,r}\big]\\
&\le \rho,
\end{align*}
where the last inequality follows from $\rho$-pointwise replicability.

By Markov's inequality, for every $\tau>0$,
\[
\Pr_{S\sim \mathcal D^n}\!\big[\Delta(S)>\tau\big]
\le \frac{\mathbb E_{S\sim \mathcal D^n}[\Delta(S)]}{\tau}
\le \frac{\rho}{\tau}.
\]

\end{proof}

\restateFewWrongRows*
\begin{proof}
A row $r$ has $v_r\neq f^*(x)$ only if the incorrect bit occurs in more than half of the probability mass of that row, i.e., only if $u_r>\tfrac12$.
By Markov's inequality,
\[
\Pr_{r}\!\big[u_r>\tfrac12\big] \;\le\; \frac{\E_r[u_r]}{1/2} \;=\; 2q_x.
\]
\end{proof}

\restateCorMonoError*
\begin{proof}
If $S$ is not $\tau$-bad, then $\Delta(S)\le\tau$. For a fresh seed $r$,
\[
\{\mathcal A(S;r)(x)\ne f^*(x)\}
\subseteq
\{v_r\ne f^*(x)\}\cup\{\mathcal A(S;r)(x)\ne v_r\}.
\]
Hence, the single-run error over $r$ is at most $m+\Delta(S)\le m+\tau$. Since the meta-algorithm uses $k$ independent fresh seeds, the conjunctive error is at most $(m+\tau)^k$.
\end{proof}

\section{Shared Pretrained Components}
\label{app:shared-pretrained-components}

Our framework focuses on models sharing training data. We briefly discuss how sharing a pretrained backbone can introduce an additional source of dependence between their errors.

A fixed pretrained backbone $B$ can be absorbed into the learning algorithm by defining $A_B(S;r)$, so our results apply unchanged whenever the required accuracy and pointwise-replicability conditions hold for the fine-tuning procedure with $B$ fixed.

A random backbone shared by all $k$ models introduces an additional shared source of error. This can be formalized by conditioning on $B$. For example, suppose the fine-tuning stages are independent conditional on $B$, and let
\[
q_{x,B}
=
\Pr_{S,r}\!\left[
A_B(S;r)(x)\neq f^*(x)\mid B
\right].
\]
The conjunctive error with a shared backbone is then $\mathbb{E}_B[q_{x,B}^k]$, whereas independently trained backbones would give $(\mathbb{E}_B[q_{x,B}])^k$. By Jensen's inequality,
\[
\mathbb{E}_B[q_{x,B}^k]
\geq
\bigl(\mathbb{E}_B[q_{x,B}]\bigr)^k.
\]

Thus, variability in backbone-specific error can increase the conjunctive error when the backbone is shared. If the fine-tuning sample is also shared, the same argument conditions jointly on $B$ and $S$. A backbone that creates a shared blind spot cannot generally be repaired by rerunning only the fine-tuning stage. In contrast, if the conditional learner remains accurate and pointwise replicable, our same-sample result can be applied conditional on the backbone.

\newpage

\end{document}